\documentclass[letterpaper, 10 pt, conference]{ieeeconf}  

\usepackage{graphics} 
\usepackage{epsfig} 
\usepackage{times} 
\usepackage{amsmath} 
\usepackage{amssymb}  

\usepackage{adjustbox}
\usepackage{amsthm}
\usepackage{amsfonts}
\usepackage{empheq}
\newtheorem{proposition}{Proposition}
\usepackage{subfigure}
\usepackage{booktabs}
\usepackage{algorithm}
\usepackage{algpseudocode} 
\usepackage{tabularx}
\usepackage{booktabs}
\usepackage{multirow}
\usepackage{array}
\usepackage{graphicx}
\usepackage{cite}

\title{\LARGE \bf STORM: Spatial-Temporal Iterative Optimization for Reliable Multicopter Trajectory Generation}

\author{Jinhao Zhang$\textsuperscript{‡}$, Zhexuan Zhou$\textsuperscript{‡}$, Wenlong Xia, Youmin Gong and Jie Mei$^{*}$
\thanks{‡ J. Zhang and Z. Zhou contribute equally to this work.}
\thanks{* Corresponding Author.}
\thanks{J. Zhang, Z. Zhou, W. Xia, Y. Gong and J. Mei are with the Department of Automation, School of Intelligence Science and Engineering, Harbin Institute of Technology, Shenzhen, China. For correspondence:{\tt\small jmei@hit.edu.cn}}
}

\begin{document}

\maketitle
\thispagestyle{empty}
\pagestyle{empty}

\begin{abstract}
Efficient and safe trajectory planning plays a critical role in the application of quadrotor unmanned aerial vehicles. Currently, the inherent trade-off between constraint compliance and computational efficiency enhancement in UAV trajectory optimization problems has not been sufficiently addressed. To enhance the performance of UAV trajectory optimization, we propose a spatial-temporal iterative optimization framework. Firstly, B-splines are utilized to represent UAV trajectories, with rigorous safety assurance achieved through strict enforcement of constraints on control points. Subsequently, a set of QP-LP subproblems via spatial-temporal decoupling and constraint linearization is derived. Finally, an iterative optimization strategy incorporating guidance gradients is employed to obtain high-performance UAV trajectories in different scenarios. Both simulation and real-world experimental results validate the efficiency and high-performance of the proposed optimization framework in generating safe and fast trajectories. Our source codes will be released for community reference.\footnote[1]{https://hitsz-mas.github.io/STORM}

\end{abstract}

\section{INTRODUCTION}

Unmanned Aerial Vehicles (UAVs) are becoming increasingly prevalent, especially with the development of low-altitude economy, such as industrial inspections and package delivery \cite{hanover2024autonomous}. For UAV, generating high-quality trajectories and ensuring safety are of critical importance. UAV rely on excellent and efficient planners to generate safe and agile trajectories, enabling reliable navigation in complex environments.

In the domain of UAV trajectory generation, the simultaneous assurance of rigorous safety constraints and computationally efficient generation of high-performance trajectories compliant with dynamic constraints remains a critical challenge. While some existing methods have achieved progress in trajectory generation, their safety verification mechanisms exhibit insufficient assurance of safety compliance\cite{8206119}. Current approaches relying on discrete sampling points for feasibility checks can only provide probabilistic safety guarantees, failing to mathematically ensure full-state safety across the continuous configuration space. This discrete verification paradigm introduces inherent risks of inter-sample collision vulnerabilities. More critically, prevalent trajectory generation algorithms often rely on conservative approximations when handling UAV dynamic constraints. Such incomplete constraint satisfaction not only restricts the full exploitation of UAV maneuverability but also induces stability deterioration during trajectory tracking, thereby compounding safety hazards.

\begin{figure}
    \centering
    \includegraphics[width=0.8\linewidth]{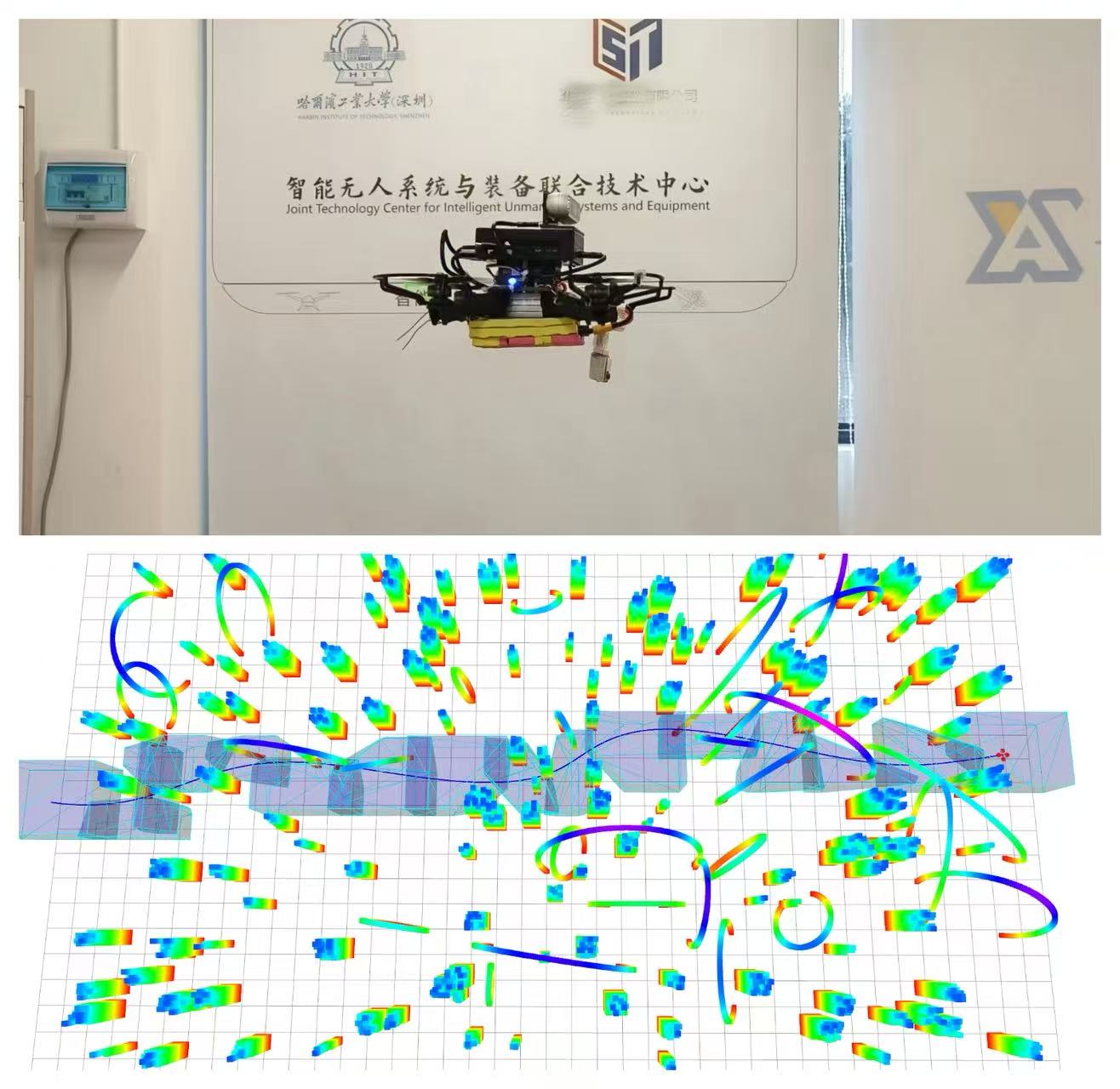}
    \caption{Real UAV Flight and Simulation}
\end{figure}

In this paper, we propose a non-uniform B-spline trajectory generation method based on spatial-temporal iterative optimization, as illustrated in Fig. 2. The proposed framework leverages the convex hull property of B-splines to transform collision constraints into sparse control point constraints, formulating a spatial optimization problem which rigorously ensuring trajectory safety. During temporal optimization of which constraints are enforced through neighborhood-based linear approximations, systematic parameter are tuned to achieve an optimal balance between dynamic performance and computational efficiency. To address the optimization problem, an efficient iterative optimization algorithm is proposed. The contributions of this paper are summarized as follows:
\begin{itemize}
\item A spatial-temporally decoupled optimization framework is proposed, which maximizes compliance with safety and dynamic constraints while achieving enhanced computational efficiency through iterative optimization mechanisms.
\item A trajectory generation method based on a safe flight corridor, leveraging the proposed optimization framework to achieve agile flight in complex environments. The proposed method enables tunable generation of trajectories compliant with diverse performance metrics through parametric adjustment mechanisms.
\item A series of comparative simulations and experimental studies are conducted, demonstrating superior performance of proposed method compared with state-of-the-art counterparts. 
\end{itemize}

\begin{figure*}[htbp]
\centering
\includegraphics[width=18cm]{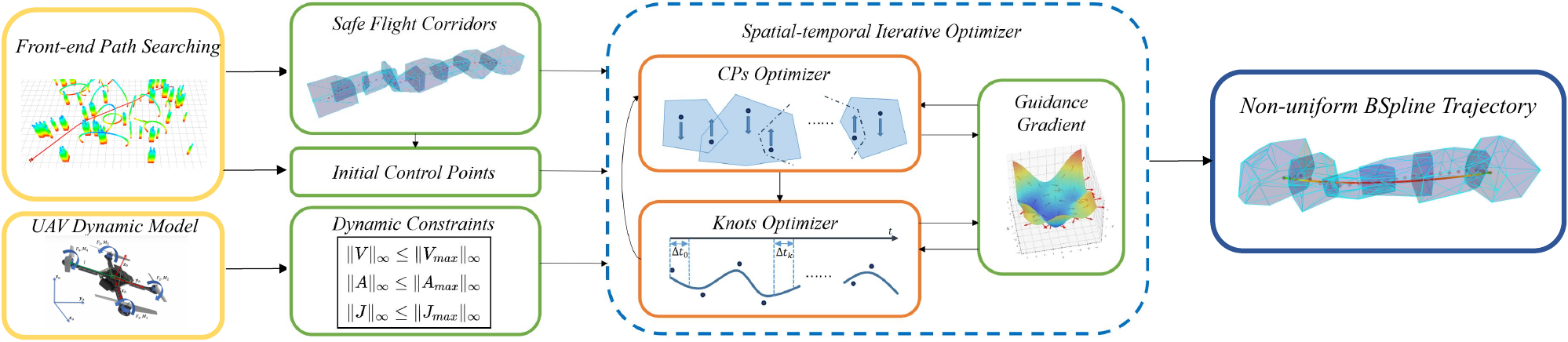}
\caption{The trajectory generation procedure of proposed method. The optimizer acquires initial control points and safe flight corridor constraints from front-end path searching results, derives dynamic constraints from the UAV dynamic model, and iteratively generates non-uniform B-spline trajectory.}
\label{fig:picture001}
\end{figure*}

\section{RELATED WORKS}

Motion planning is a critical high-level module that coordinates the navigation UAVs in complex environments. The motion planning problem is essentially formulated as a constrained optimization problem, where the efficiency and feasibility of the solution are of utmost importance. The prevailing trajectory optimization approaches are broadly categorized into two paradigms: hard-constrained methods that strictly enforce feasibility through explicit constraints, and soft-constrained approaches that incorporate constraint violations as penalty terms within the optimization objective.

The pioneering work in hard constraint methods is minimum snap trajectory generation \cite{mellinger2011minimum}, which generates piecewise polynomials and solves the optimized trajectory via quadratic programming (QP). \cite{richter2016polynomial} proposes an optimized iterative time allocation algorithm that can configure the time of each polynomial segment to satisfy the kinematic constraints of the UAV. \cite{ding2018trajectory} proposes a B-spline based kinodynamic search algorithm to find an initial trajectory, which is then refined by an elastic band optimization approach. The convex hull properties of B-splines can ensure geometric constraints but fails to fully exploit the dynamic performance of the drone. The trajectory time allocation in the above methods relies on simplistic heuristics, whose suboptimality may result in low-quality trajectories. 

To solve these problems, \cite{gao2018online} proposes a method to generate online safe and dynamically feasible trajectories for quadrotors by optimizing the search for paths with a reasonable time allocation. \cite{sun2021fast} separates state variables from time variables to decompose the nonlinear programming (NLP) trajectory optimization problem. A series of simple convex regions along a rough trajectory in the environment is generated to form safe flight corridors, in which the trajectories are restricted \cite{chen2015real,chen2016online,liu2017planning,gao2020teach}. \cite{zhou2023efficient} introduces an innovative method for calculating time-optimal trajectories by anchoring waypoints through specific constraints and employing distinct sampling intervals between each pair of waypoints, thereby substantially enhancing the efficiency of trajectory planning.


Another alternative is to express the kinematic and dynamic constraints of trajectory optimization as a soft constraint problem, considering both smoothness and safety. \cite{zucker2013chomp} generates discrete-time trajectories by minimizing smoothness and collision costs through gradient-based methods, while integrating sampling-based or graph-based search strategies within the framework. \cite{zhou2019robust} proposes a hierarchical optimization framework using B-spline curves, combining front-end path search with back-end trajectory optimization. The dynamic constraints and obstacle-avoidance constraints are incorporated into the optimization problem, enabling real-time planning in complex environments. Ego-planner \cite{zhou2020ego} is an ESDF-free gradient-based local replanning framework, supporting multi-agent collaboration and dynamic obstacle avoidance. \cite{ren2022bubble} uses a series of overlapping spheres to represent the free space of the environment for flight corridor planning. As the mainstream solution in the current planning field, MINCO \cite{wang2022geometrically} leverages the differential flatness property of quadrotors to eliminate dynamics constraints. And it uses diffeomorphic transformations and discrete point sampling to convert geometric constraints into unconstrained optimization problems. 

The optimization of hard-constrained problems incurs prohibitive computational costs and may fails to satisfy real-time requirements. Furthermore, overly conservative kinodynamic constraints produce trajectory velocities insufficient to support aggressive flight maneuvers. Concurrently, soft-constrained methods may not be able to rigorously enforce full-trajectory geometric feasibility or kinodynamic compliance, thereby severely limiting their practical applicability in some real-world scenarios. Comprehensively considering the aforementioned issues, this paper proposes a method that synergistically integrates the merits of both aforementioned approaches and establishes a superior framework for UAV trajectory optimization.

\section{PROBLEM FORMULATION}
Our research objective is to devise a parameterized trajectory function represented by $C(t)$: $[t_s, t_f] \rightarrow \mathbb{R}^3$ that simultaneously satisfies multiple critical criteria: temporal optimality through total duration minimization, spatial optimality via jerk integral minimization, along with safety guarantees, kinodynamic feasibility, and boundary state constraints. The proposed method utilizes B-spline parameterization, strategically leveraging its inherent convex hull property to efficiently enforce safety boundaries and kinematic constraints. To enable adaptive spatial-temporal deformation capabilities, the conventional B-spline trajectory generation framework is extended by incorporating non-uniform temporal knot spacing, thereby establishing a unified spatial-temporal optimization paradigm. 

\begin{algorithm} [t]
\caption{Spatial-Temporal Iterative Optimization.}\label{alg:iopt}
\begin{algorithmic}[1] 
\Require  $\mathbf{Q}_0$, $\mathbf{T}_0$; $\mathbf{G}$, $h$, $p_s$, $p_f$, $v_s$, $v_f$; $\rho$, $\rho_m$, $\epsilon$ 
\Ensure Optimal trajectory represented by $\mathbf{Q}^*$, $\mathbf{T}^*$ 

\State {\bf Initialize}$()$
\State $\mathbf{Q}^* \gets \textbf{None} $, $\mathbf{T}^* \gets \textbf{None}$
\For{$k = 1$ to $M$}
    \If{$\rho \leq \rho_m$} 
    \State{\bf UpdateGuidance}$(\mathbf{Q}_k$, $\mathbf{T}_k)$
    \State $g_1, g_2 \gets$ {\bf GetGuidance}$(c)$
    \State {\bf AddGuidance}$(g_1, g_2)$
    \EndIf
    \State $\mathbf{Q}_k \gets$ {\bf OptimizeCPs}$()$
    \State {\bf UpdateCPs}$(\mathbf{Q}_k)$
    \State $\mathbf{T}_k \gets$ {\bf OptimizeKnots}$()$
    \If{$\mathbf{T}_k$ is {\bf None} and $k>m$} 
        \State $\mathbf{Q}^* \gets \mathbf{Q}_{k-1} $, $\mathbf{T}^* \gets \mathbf{T}_{k-1} $
        \State \Return $\mathbf{Q}^*$, $\mathbf{T}^*$
    \EndIf
    \If{{\bf sum}($|\mathbf{T}_{k}-\mathbf{T}_{k-1}|)<\epsilon$}
        \State $\mathbf{Q}^* \gets \mathbf{Q}_{k} $, $\mathbf{T}^* \gets \mathbf{T}_{k} $
        \State \Return $\mathbf{Q}^*$, $\mathbf{T}^*$
    \EndIf
    \State {\bf UpdateKnots}$(\mathbf{T}_k)$
    \State {\bf UpdateDecayFactor}$(k)$
\EndFor
\State $\mathbf{Q}^* \gets \mathbf{Q}_{M} $, $\mathbf{T}^* \gets \mathbf{T}_{M} $
\State \Return $\mathbf{Q}^*$, $\mathbf{T}^*$
\end{algorithmic}
\end{algorithm}

Normally, let $\mathbf{Q}=[Q_0^{\mathrm{T}},\dots,Q_n^{\mathrm{T}}]^{\mathrm{T}} \in \mathbb{R}^{3(n+1)}$ denote the control points and $[t_0,t_1,\dots t_{n+p}]^{\mathrm{T}}$ denote the knots vector. Then, the knots span is given by $\mathbf{T}=[\Delta t_0,\Delta t_1,\dots \Delta t_m]^{\mathrm{T}}$, where $\Delta t_i = t_{i+1}-t_{i}$.
Our objective optimization problem is expressed as follows
\begin{align}
    \min_{\mathbf{Q},\mathbf{T}}  & \;\mathcal{J}= \frac{1}{2}\; \int_{t_s}^{t_f}\|J_{\scriptscriptstyle{\mathbf{Q},\mathbf{T}}}\|^2{\rm d}t + \rho \mathbf{1}^{\mathrm{T}} \mathbf{T}, \label{eq1}\\
    {\rm s.t. } & \; C(t) \subseteq \mathcal{F}, \forall t \in [t_s, t_f], \label{eq2}\\
    & \; C(t_s)=p_s,\; C(t_f)=p_f, \label{eq3}\\
    & \; \dot{C}(t_s)=v_s,\; \dot{C}(t_f)=v_f, \label{eq4}\\
    & \; \|\mathbf{V}\|_{\infty}  \leq V_{\max},\;\|\mathbf{A}\|_{\infty}  \leq A_{\max} ,\;\|\mathbf{J}\|_{\infty}  \leq J_{\max}, \label{eq5} \\
    & \; \mathbf{T} \succeq 0, \label{eq6}
\end{align}
where $J_{\scriptscriptstyle{\mathbf{Q},\mathbf{T}}}$ is the jerk of the trajectory, $\mathcal{F}$ denotes the Safe Flight Corridors (SFCs) constructed by the front-end discrete path searching, $\rho$ denotes temporal weight, while $\mathbf{V} \in \mathbb{R}^{3n}$, $\mathbf{A} \in \mathbb{R}^{3(n-1)}$ and $\mathbf{J} \in \mathbb{R}^{3(n-2)}$ representing the velocity, acceleration, and jerk control points of the motion trajectory, respectively. Moreover, to naturally satisfy the boundary conditions, a quasi-uniform B-spline inspired approach is adopted: the first and last knots are repeated $p$ times ($p$ being the B-spline degree). This configuration ensures that the terminal control points directly correspond to the start and end points of the spline curve.

\section{Spatial-Temporal Iterative Optimization}
In this section, a novel iterative optimization framework is proposed to address the problems \eqref{eq1}--\eqref{eq6}. First, the matrix representation of B-splines is introduced to facilitate the formulation of subsequent optimization problems (Section \ref{mr}). Subsequently, the original problem is decoupled into temporal and spatial dimensions (Section \ref{std}). Then, the constraints are linearized to enhance computational efficiency (Section \ref{cl}). To compensate for the loss of temporal weight information caused by the spatial-temporal decoupling, guidance gradients are employed to coordinate the updates of both optimizers (Section \ref{4.3}). Finally, the optimal solution to the original problem is obtained by iteratively solving simplified sub-problems and the algorithmic procedure is summarized (Section \ref{iof}). The complete algorithmic process is detailed in Algorithm \ref{alg:iopt}.

\subsection{Matrix Representations for Non-Uniform B-Splines} \label{mr}
The recursive computation method proposed in \cite{qin1998general} is utilized to derive the matrix representation of non-uniform B-splines. 

Consider a B-spline trajectory $C(t)$ of degree $p$, the $i$-th segment of the B-spline trajectory, $C_{i-p}(u)$, is given by
\begin{align}
    C(t)=C_{i-p}(u)=\begin{bmatrix}
    1, u, \dots, u^{p-1}
\end{bmatrix} \mathbf{M}^{p+1}_i\begin{bmatrix}
    Q_{i-p} \\
    Q_{i-p+1} \\
     \vdots \\
     Q_i
\end{bmatrix},
\end{align}
where $u=\frac{t-t_i}{t_{i+1}-t_i}, t \in [t_i, t_{i+1})$, and $\mathbf{M}^{p+1}_i$ is the $i$-th basic matrix of B-spline basis functions of degree $p$. To facilitate a more convenient representation of nonuniform B-splines, the following proposition is adopted. 
\begin{proposition} 
Let $\mathbf{M}^1_i = \begin{bmatrix} 1 \end{bmatrix}$, then the basic matrix can be recursively calculated as follows
\begin{align}
    &\mathbf{M}^{p+1}_i =    
    \begin{bmatrix} 
        \mathbf{M}^{p}_i \\ 
        \mathbf{0}
    \end{bmatrix} 
    \begin{bmatrix}
        1-d^0_{i-p+1} & d^0_{i-p+1}       &           & 0         \\
                      & \ddots            & \ddots    &           \\
        0             &                   &1-d^0_{i}  & d^0_{i}
    \end{bmatrix} \notag
    \\[1.5ex]
    &\phantom{\mathbf{M}^{p+1}_i} + 
    \begin{bmatrix}
        \mathbf{0} \\ 
        \mathbf{M}^{p}_i
    \end{bmatrix}
    \begin{bmatrix}
        -d^1_{i-p+1} & d^1_{i-p+1}      &           & 0         \\
                     & \ddots           & \ddots    &           \\
        0            &                  &-d^1_{i}   & d^1_{i}
    \end{bmatrix}, 
\end{align}
where $d^0_j=\frac{t_i-t_j}{t_{j+p}-t_j}$ and $ d^1_j=\frac{t_{i+1}-t_i}{t_{j+p}-t_j}$, with the convention $0/0=0$.
\end{proposition}
By leveraging this proposition, each point on the B-spline can be expressed as a linear combination of the control points, thereby facilitating the formulation of subsequent optimization problems. For further details, we recommend that readers refer to the original article \cite{qin1998general}.

\subsection{Spatial-Temporal Decoupling} \label{std}
The objective function \eqref{eq1} is highly non-convex, making it extremely challenging to solve and computationally intensive. A more effective approach is to decouple it into simpler, more tractable sub-problems. Thereby, the original problem is decoupled into two components: spatial optimization with safety and boundary constraints, and temporal optimization under kinodynamic constraints
\begin{align}
    &\left\{
    \begin{aligned}
        \min_{\mathbf{Q}} \; & \frac{1}{2}\; \int_{t_s}^{t_f}\|J_{\scriptscriptstyle{\mathbf{Q},\mathbf{T}}}\|^2{\rm d}t,  \\
        {\rm s.t.} \; & C(t) \subseteq \mathcal{F}, \forall t \in [t_s, t_f] ,\\
         \; & C(t_s)=p_s,\; C(t_f)=p_f,  \\
         \; & \dot{C}(t_s)=v_s,\; \dot{C}(t_f)=v_f, 
    \end{aligned} 
    \right.   \\ \notag  \\ 
    &\left\{
    \begin{aligned}
        \min_{\mathbf{T}} \; & \mathbf{1}^{\mathrm{T}} \mathbf{T},  \\
        {\rm s.t. }\; & \|\mathbf{V}\|_{\infty}  \leq V_{\max}, \\ \;&\|\mathbf{A}\|_{\infty}  \leq A_{\max} ,\\ \;&\|\mathbf{J}\|_{\infty}  \leq J_{\max}, \\
        \; & \; \mathbf{T} \succeq 0.
    \end{aligned} 
    \right. 
\end{align}

Note that each optimizer treats the other variables as fixed, optimizing the control points $\mathbf{Q}$ and the knots span $\mathbf{T}$ separately. When $\mathbf{Q}$ is fixed, the temporal term in objective function \eqref{eq1} is linear with respect to $\mathbf{T}$.
When $\mathbf{T}$ is fixed, the energy term in the objective function \eqref{eq1} is quadratic with respect to $\mathbf{Q}$
\begin{align}
   & \int_{t_s}^{t_f}\|J_{\scriptscriptstyle{\mathbf{Q},\mathbf{T}}}\|^2{\rm d}t \\
    =& \sum_{i}\int_{t_i}^{t_{i+1}}\|J_{\scriptscriptstyle{\mathbf{Q},\mathbf{T}}} \|^2 {\rm d}t \\
    =& \sum_{i} \Delta t_i  \int_{0}^{1} \|J_{\scriptscriptstyle{\mathbf{Q},\mathbf{T}}}(u) \|^2  {\rm d}u \\
    =& \sum_{i} J^{[i],\mathrm{T}}\mathbf{P}_i J^{[i]}\Delta t_i  \\
    =& \sum_{i} Q^{[i],\mathrm{T}}\mathbf{W}_i Q^{[i]}\Delta t_i ,
\end{align}
where
\begin{equation}
    Q^{[i]}=[Q_{i-p+3}^{\mathrm{T}}, \dots, Q_{i+3}^{\mathrm{T}}]^{\mathrm{T}},
\end{equation}
\begin{equation}
    J^{[i]}=[J_{i-p+3}^{\mathrm{T}}, \dots, J_{i}^{\mathrm{T}}]^{\mathrm{T}}.
\end{equation}
Let $J^{[i]}=\mathbf{L}^{[i]}Q^{[i]}$, $\mathbf{U}=[1, \dots, u^{p-3}]^{\mathrm{T}}$, then
\begin{equation}
    \mathbf{P}_i = \int_{0}^1 \mathbf{M}^{p-3,\mathrm{T}}_i \mathbf{U}\mathbf{U}^{\mathrm{T}} \mathbf{M}^{p-3}_i {\rm d}u,
\end{equation}
\begin{equation}
    \mathbf{W}_i = \mathbf{L}^{[i], \mathrm{T}}\mathbf{P}_i\mathbf{L}^{[i]}.
\end{equation}

Nevertheless, such decoupling leads to the loss of temporal weight information. Therefore, guidane gradient from the original objective function is introduced to direct the iterative updates, as detailed in Section \ref{4.3}.

\subsection{Constraint Handling} \label{cl}

\subsubsection{Spatial Constraints}
Owing to the fact that SFCs are formed by unions of convex polyhedrons or spheres, the safety constraints \eqref{eq2} in the spatial optimization problem are intrinsically linear, which can be expressed as
\begin{equation}
\begin{bmatrix}
G_0 & 0 & \cdots & 0 \\
0 & G_1 & \cdots & 0 \\
\vdots & \vdots & \ddots & \vdots \\
0 & 0 & \cdots & G_n \\
\end{bmatrix} 
\begin{bmatrix}
Q_0\\Q_1\\ \vdots \\Q_n
\end{bmatrix} \preceq
\begin{bmatrix}
h_0\\h_1\\ \vdots \\h_n
\end{bmatrix}.
\end{equation}
For a simplified notation, it can be represented as
\begin{equation}
    \mathbf{G}\mathbf{Q}\preceq h. \label{c11}
\end{equation}

Additionally, with fixed knots span, the velocity control points $\mathbf{V}=[V_0^{\mathrm{T}}, \dots, V_{n-1}^{\mathrm{T}}]^{\mathrm{T}}$ can be obtained from the trajectory control points $\mathbf{Q}$ via the following linear transformation
\begin{equation}
    V_i = \frac{p(Q_{i+1}-Q_i)}{\Delta t_{i-p+1}+\dots+\Delta t_i}, \label{eq4.2.3}
\end{equation}
where $\Delta t_i$, for $i <0$ or $i > m$, is set to be $0$ here. Leveraging the matrix formulation of non-uniform B-splines, we can conveniently convert the boundary conditions \eqref{eq3}--\eqref{eq4} into linear constraints
\begin{equation}
    \mathbf{A}\mathbf{Q}=b. \label{c12}
\end{equation}

\subsubsection{Temporal Constraints}
The control points for acceleration and jerk of the trajectory can be computed as
\begin{align}
    A_i &= \frac{(p-1)(V_{i+1}-V_i)}{\Delta t_{i-p+2}+\dots+\Delta t_i}, \label{eq4.2.5} \\
    J_i &= \frac{(p-2)(A_{i+1}-A_i)}{\Delta t_{i-p+3}+\dots+\Delta t_i}. \label{eq4.2.6}  
\end{align}

Define $T^{[r,s]}=[\Delta t_{r}, \dots, \Delta t_s]^{\mathrm{T}}$, $r \leq s$. Specifically, $T^{[i-p+1,i]}$, $T^{[i-p+1,i+1]}$, and $T^{[i-p+1,i+2]}$ are denoted as $T^{[i,V]}$, $T^{[i,A]}$, and $T^{[i,J]}$, respectively. Since the control points are fixed, the velocity constraint is linear with respect to $\mathbf{T}$, which can be expressed as
\begin{equation}
    \mathbf{1}^{\mathrm{T}} T^{[i, V]} \geq \frac{p}{V_{\max}} \| Q_{i+1} - Q_i \|_{\infty} . \label{eq4.3.1}
\end{equation}

However, the velocity and acceleration control points exhibit a nonlinear dependency on $\mathbf{T}$. Therefore, a linear approximation strategy is adopted: the value of $V_i$ and $A_i$ within a certain neighborhood of $\mathbf{T}_k$ is directly approximated by $V_i(T_k^{[i, V]})$ and $A_i(T_k^{[i, A]})$, respectively, as follows
\begin{equation}
    V_i(T^{[i, V]}) \approx V_i(T_k^{[i, V]}), \forall T^{[i, V]} \in \mathcal{N}(T_k^{[i, V]}),
\end{equation}
\begin{equation}
    A_i(T^{[i, A]}) \approx A_i(T_k^{[i, A]}), \forall T^{[i, A]} \in \mathcal{N}(T_k^{[i, A]}).
\end{equation}
Consequently, the constraints on acceleration and jerk are linearized as follows
\begin{equation}
    \mathbf{e}_1^{\mathrm{T}} T^{[i, A]} \geq \frac{p-1}{A_{\max}}\|V_{i+1}(T_k^{[i+1, V]})-V_i(T_k^{[i, V]})\|_{\infty}, \label{eq4.3.2}
\end{equation}
\begin{equation}
    \mathbf{e}_2^{\mathrm{T}} T^{[i, J]} \geq \frac{p-2}{J_{\max}}\|A_{i+1}(T_k^{[i+1, A]})-A_i(T_k^{[i, A]})\|_{\infty}, \label{eq4.3.3}
\end{equation}
where
\begin{equation*}
     \mathbf{e}_1 = [0, 1, \dots, 1, 0]^{\mathrm{T}} \in \mathbb{R}^{p+1} ,
\end{equation*}
\begin{equation*}
    \mathbf{e}_2 = [0, 0, 1, \dots, 1, 0, 0]^{\mathrm{T}} \in \mathbb{R}^{p+2}.
\end{equation*}

To improve the accuracy of the expansion, the feasible domain is restricted to a neighborhood of $\mathbf{T}_k$. Thus, a decay factor $0<\gamma_{i, k} < 1$ is employed to control the descent rate of $\Delta t_i$ in iterations
\begin{equation}
    \Delta t_{i, k}-\Delta  t_i \leq \gamma_{i, k}\Delta t_{i, k},
\end{equation}
which is equivalent to
\begin{equation}
    \Delta t_i \geq (1-\gamma_{i, k})\Delta t_{i, k} .\label{eq4.3.4}
\end{equation}
The decay factor is monotonically decreasing in each iteration towards zero to progressively enhance linearization accuracy. Our experimental findings reveal that different initial decay factors lead to distinct performance trajectories, enabling us to select the most appropriate initial decay factor based on specific requirements.

Additionally, to ensure that the total time decreases monotonically, the total time from the previous iteration is used as the upper bound for the current iteration
\begin{equation}
    \mathbf{1}^{\mathrm{T}} \mathbf{T} \leq t_{k, \max}, \label{eq4.3.5}
\end{equation}
\begin{equation}
    t_{k, \max} = \mathbf{1}^{\mathrm{T}} \mathbf{T}_k.
\end{equation}
Specifically, in the first iteration, $t_{0, \max}$ is set to be the total length of the polyline formed by the front-end waypoints.

\begin{algorithm}
\caption{UpdateGuidance and GetGuidance}\label{alg:ggu}
\begin{algorithmic}[1]
    \Function{UpdateGuidance}{$\mathbf{Q}$, $\mathbf{T}$}
        \State $\text{grad}_{\text{new}} \gets \nabla_{\mathbf{Q},\mathbf{T}} \mathcal{J}$
        \State $\text{grad} \gets \gamma \cdot \text{grad} + \text{grad}_{\text{new}}$
        \State $\text{norm\_grad} \gets \text{grad} / \|\text{grad}\|_2$
    \EndFunction

    \State

    \Function{GetGuidance}{$c, \rho$}
        \State $\text{grad}_{\mathbf{Q}} \gets c \cdot \text{norm\_grad}_{\mathbf{Q}}$
        \State $\text{grad}_{\mathbf{T}} \gets \frac{c}{\rho} \cdot \text{norm\_grad}_{\mathbf{T}}$
        \State \Return $\text{grad}_{\mathbf{Q}}$, $\text{grad}_{\mathbf{T}}$
    \EndFunction
\end{algorithmic}
\end{algorithm}

\subsection{Guidance Gradient}\label{4.3}
To address the temporal information loss in weighting induced by decoupling, the iterative processes of dual optimizers are synchronized using guidance gradients derived from the primary objective function \eqref{eq1}. This enables updating the steering parameter along the descent directions of $\mathcal{J}$.

Formally, suppose the normalized gradient of the original objective function $\mathcal{J}$ is
\begin{equation}
    \nabla_{\mathbf{Q},\mathbf{T}} \mathcal{J}=\begin{bmatrix}
        \nabla_{\mathbf{Q}} \mathcal{J}_{\mathbf{T}} \\
        \nabla_{\mathbf{T}} \mathcal{J}_{\mathbf{Q}} 
    \end{bmatrix}. \label{ng}
\end{equation}
Instead of directly leveraging the gradient in the optimization process, the gradient information is strategically incorporated by augmenting the objective function with a linear gradient-augmented term
\begin{align}
    &\frac{1}{2}\; \int_{t_s}^{t_f}\|J_{\scriptscriptstyle{\mathbf{Q},\mathbf{T}}}\|^2{\rm d}t + \rho \mathbf{1}^{\mathrm{T}} \mathbf{T} + c\nabla_{\mathbf{Q},\mathbf{T}} \mathcal{J}^{\mathrm{T}}\begin{bmatrix}
        \mathbf{Q}\\
        \mathbf{T}
    \end{bmatrix} \\
     = & \underbrace{\frac{1}{2} \int_{t_s}^{t_f}\|J_{\scriptscriptstyle{\mathbf{Q},\mathbf{T}}}\|^2{\rm d}t + c(\nabla_{\mathbf{Q}}\mathcal{J}_{\mathbf{T}})^{\mathrm{T}}\mathbf{Q} }_{\rm Spatial}
    + \rho(\underbrace{\mathbf{1} + \frac{c}{\rho}\nabla_{\mathbf{T}} \mathcal{J}_{\mathbf{Q}})^{\mathrm{T}}\mathbf{T}}_{\rm Temporal},\notag
\end{align}
where $c \in \mathbb{R}_+$ is the confidence level of the guidance gradient. 

When $\rho$ is sufficiently large ($\rho\geq 100$ in our experiments), the terms related to $\mathbf{Q}$ in the normalized gradient \eqref{ng} all become zero, while the terms associated with $\mathbf{T}$ all converge to one. In this case, the gradient-augmented term does not alter the formulation of the decoupled sub-problems. To simplify the computation, a threshold $\rho_m$ is introduced. When $\rho > \rho_m$, the gradient augmentation term can be omitted without affecting the collaborative updating of the decoupled sub-problems. Iterations without the gradient-augmented term tend to prioritize updates aimed at minimizing the objective function those with a sufficiently large temporal weight $\rho$, which corresponds to minimizing the total duration.  

Considering the non-convex characteristics of the objective function $\mathcal{J}$, the momentum optimization approach \cite{sutskever2013importance} is implemented to update the guidance gradient
\begin{equation}
    g \leftarrow \gamma g+\nabla_{\mathbf{Q},\mathbf{T}} \mathcal{J},
\end{equation}
where $0 \leq \gamma \leq 1$ indicates the influence of historical gradients. This approach effectively mitigates gradient oscillations and accelerates the convergence rate in non-convex optimization problems, as demonstrated in \cite{sutskever2013importance}. The computation and update process of the guidance gradients is illustrated in Algorithm \ref{alg:ggu}.

Intuitively, the solution of sub-problems inherently depends on the gradient-based updates (parameter adjustment along the negative gradient direction). The proposed gradient augmentation mechanism superimposes the gradient components derived from the primal objective function into sub-problems. This synergistic modification creates the coordinated optimization trajectories, ensuring that the solutions to both sub-problems converge jointly toward configurations that effectively minimize the value of \eqref{eq1}. The results of our ablation study in Section \ref{as} validate the effectiveness of the proposed guided gradient approach.

\subsection{Iterative Optimization Framework} \label{iof}
Now, the original proble has been successfully decomposed into two components: QP and LP, with QP functioning as the control points optimizer
\begin{equation}
   \begin{aligned}
    \min_{\mathbf{Q}} \; & \sum_{i} Q^{[i],T}\mathbf{W}_i Q^{[i]}\Delta t_i + c(\nabla_{\mathbf{Q}}\mathcal{J}_{\mathbf{T}})^{\mathrm{T}}\mathbf{Q},  \\
    {\rm s.t.} \; &  \mathbf{G}\mathbf{Q} \preceq h, \mathbf{A}\mathbf{Q}=b,
\end{aligned} 
\end{equation}
and LP functioning as the knots optimizer
\begin{align}
    \min_{\mathbf{T}} \; & (\mathbf{1} + \frac{c}{\rho}\nabla_{\mathbf{T}} \mathcal{J}_{\mathbf{Q}})^{\mathrm{T}}\mathbf{T}, \\ \notag
    {\rm s.t.} \; & \mathbf{T} \succeq 0, \\ \notag
     \; & \mathbf{1}^{\mathrm{T}} \mathbf{T} \leq t_{k, \max,} \\ \notag
     \; & \Delta t_i \geq (1-\gamma_{i, k})\Delta t_{i, k},\\ \notag
    \; & \mathbf{1}^{\mathrm{T}} T^{[i, V]} \geq \frac{p}{V_{\max}} \| Q_{i+1} - Q_i \|_{\infty}, \\ \notag
     \; & \mathbf{e}_1^{\mathrm{T}} T^{[i, A]} \geq \frac{p-1}{A_{\max}}\|V_{i+1}(T_k^{[i+1, V]})-V_i(T_k^{[i, V]})\|_{\infty}, \\ \notag
     \; & \mathbf{e}_2^{\mathrm{T}} T^{[i, J]} \geq \frac{p-2}{J_{\max}}\|A_{i+1}(T_k^{[i+1, A]})-A_i(T_k^{[i, A]})\|_{\infty}.\notag
\end{align}

To obtain the optimal solution to the original problem, these two problems are solved alternately, where the guidance gradient ensures the coordinated update of both sub-problems towards minimizing the original objective function during the iterative process.

As shown in Algorithm \ref{alg:iopt}, the optimizer is initialized with quasi-uniform B-splines where the initial control points are obtained through linear interpolation of the discrete waypoints from front-end path searching. For smaller values of $\rho$, the corresponding guidance gradients are computed and integrated into the sub-problems' objective functions. Subsequently, the control points optimizer and knots optimizer are alternately solved and iterated until the total time reduction falls below $\epsilon$, which is the tolerance for total flight duration variation. Additionally, to address potential optimization failures caused by the knots optimizer overshooting the optimal solution due to excessive reduction in a given iteration, an early stopping mechanism is implemented: if the knot optimizer fails but the iteration count exceeds $m$, the algorithm returns the results from the previous iteration.

\section{EXPERIMENTS}

\subsection{Simulation Experiment Setup} 
The simulation environment incorporates complex obstacle configurations (see Fig. 5). For trajectory generation, a front-end path planner employing discrete search generates initial trajectories, which are subsequently used to construct flight corridors that enforce collision avoidance constraints in the optimization phase. During dataset generation, randomized start-goal configurations are sampled, with periodic randomized map regeneration to ensure trajectory diversity and environmental generality. B-splines of degree $p=3$ are employed, and a square-root decay strategy is uniformly applied to the decay factor $\gamma$. The temporal weight is set to $\rho=512$ (to minimize total time) and the tolerance is $\epsilon=0.05$. The experiment videos are available.\footnote[1]{https://www.bilibili.com/video/BV18Y9nYJE2z}

\subsection{Ablation Study} \label{as}
A set of ablation studies are conducted based on the aforementioned dataset comprising over 4,000 trajectories with varying lengths to evaluate the parametric sensitivity of $\textit{Decay Factor}$, $\textit{Average Energy}$, $\textit{Trajectory Length}$, and  $\textit{Average Iteration Count}$. The quantified experimental results are presented in Fig. 3.

\begin{figure}[H]
    \centering
    \includegraphics[width=0.8\linewidth]{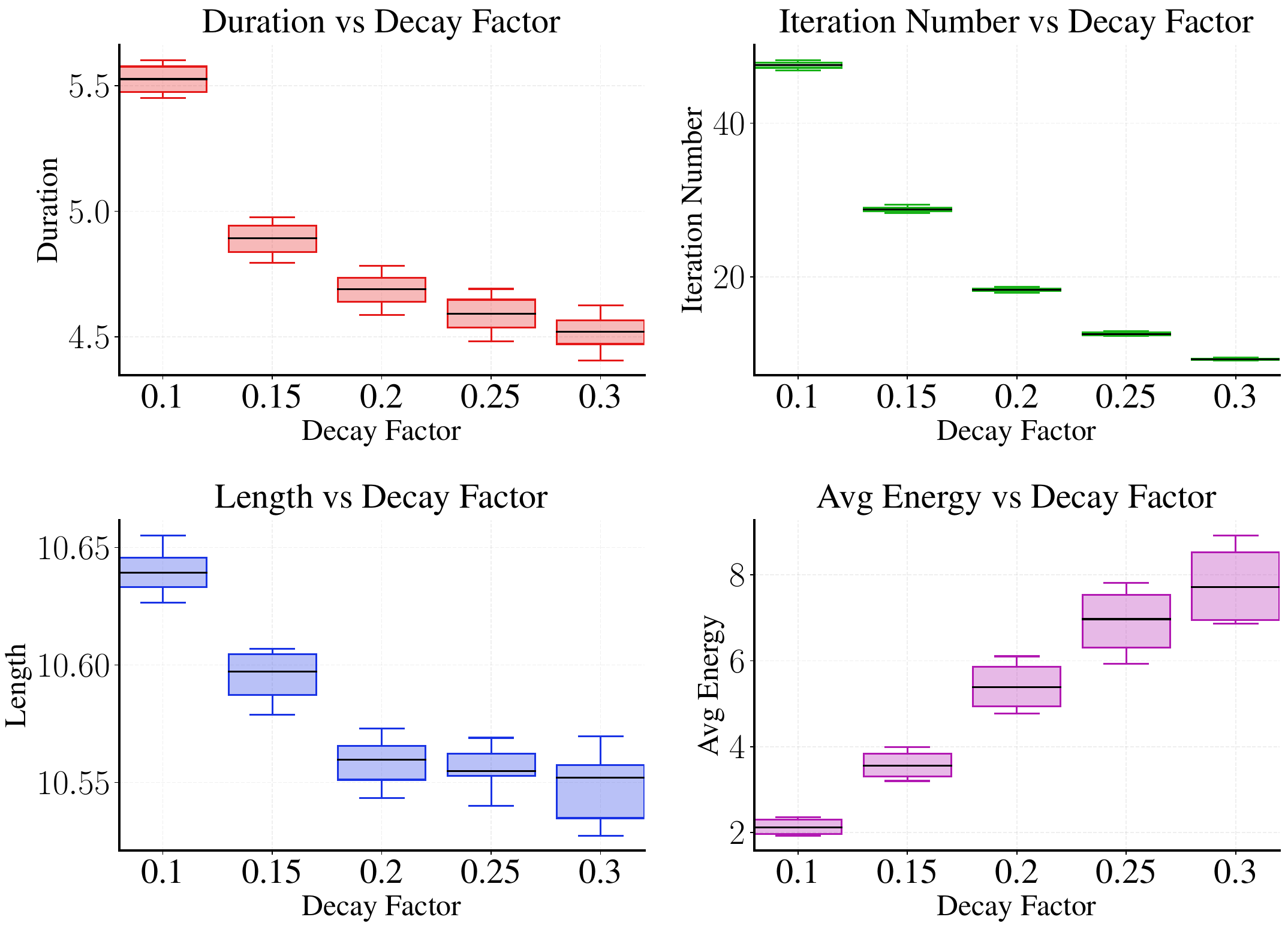}
    \caption{Ablation Experiment Results}
\end{figure}


The ablation studies reveal a critical trade-off governed by the decay factor: trajectory duration and path length exhibit monotonic reductions with increasing decay factor values, while energy expenditure rises commensurately. Concurrently, convergence iteration counts decrease across the tested decay factor range, confirming an inverse proportionality between parameter magnitude and computational overhead.

We also conduct a set of ablation experiments on the guidance gradient with different $\rho \sim [0.01, 512]$ as shown in Fig. 4. It can be observed that as $\rho$ increases (indicating greater importance of the time component), the optimal total time incorporating the guidance gradient exhibits a decreasing trend, while the energy consumption shows a corresponding increase. When $\rho \geq 100$, both metrics stabilize, suggesting that the original problem has essentially become equivalent to minimizing the total time, thereby rendering the guidance gradient unnecessary.

\begin{figure}[H]
    \centering
    \includegraphics[width=1.0\linewidth]{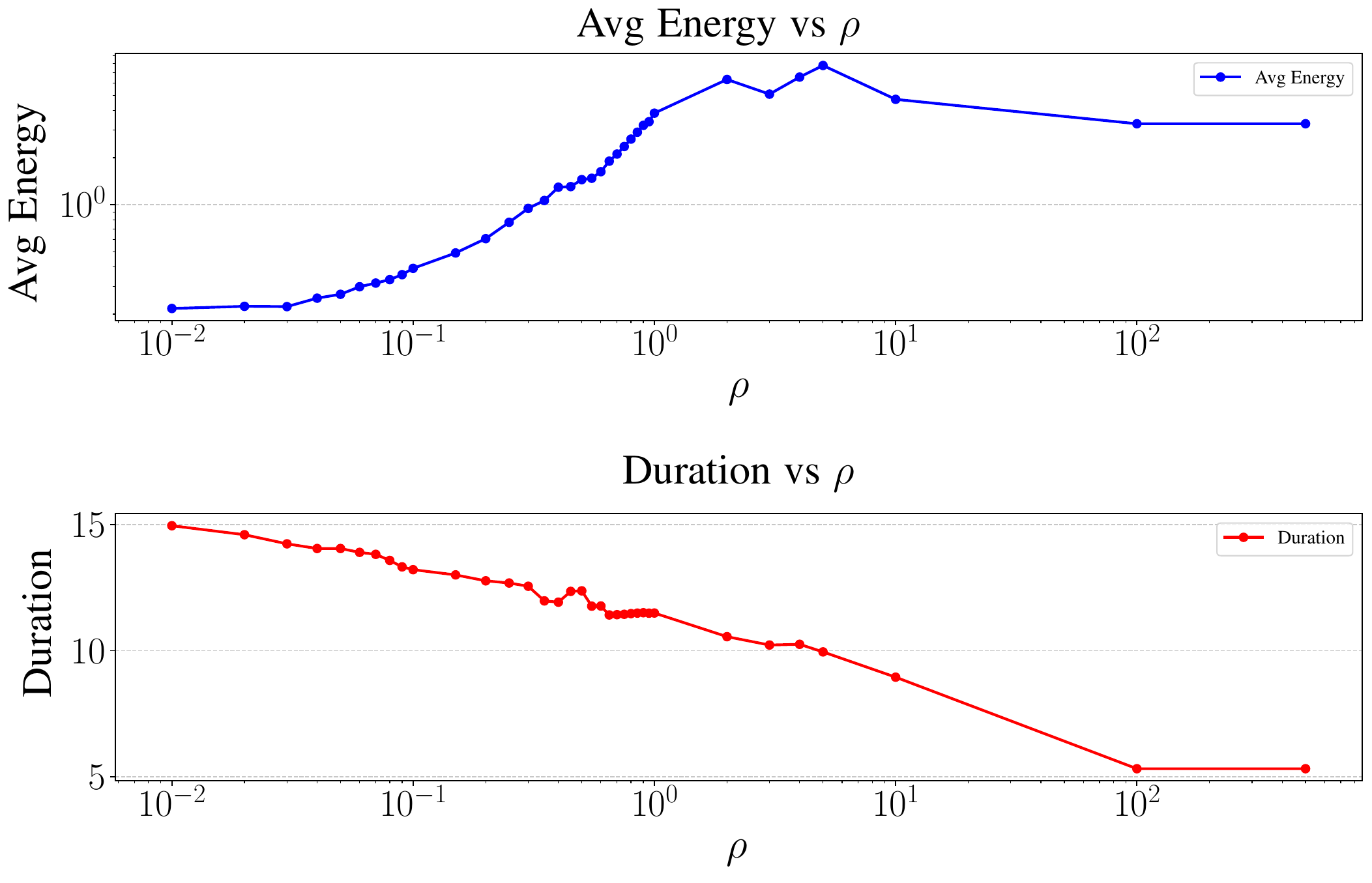}
    \caption{Effect of Guidance Gradient}
\end{figure}

\label{ablation}

\subsection{Trajectory Performance}
Given the demonstrated superiority and established adoption of MINCO trajectory \cite{wang2022geometrically} in existing researches \cite{Zhou2021DecentralizedST}, this method is selected as the comparative baseline for trajectory performance evaluation. Multi-group trajectories generated in complex environments are compared with MINCO trajectories as illustrated in Fig. 5. To ensure methodological rigor, all comparative trajectories are uniformly derived from identical front-end discrete paths and flight corridors. Also, the dynamic constraints maintain uniform throughout all experiments.

\begin{figure}[htbp]
    \centering
       \subfigure[Proposed (Blue) and MINCO (Orange)]{
        \includegraphics[width=0.23\textwidth]{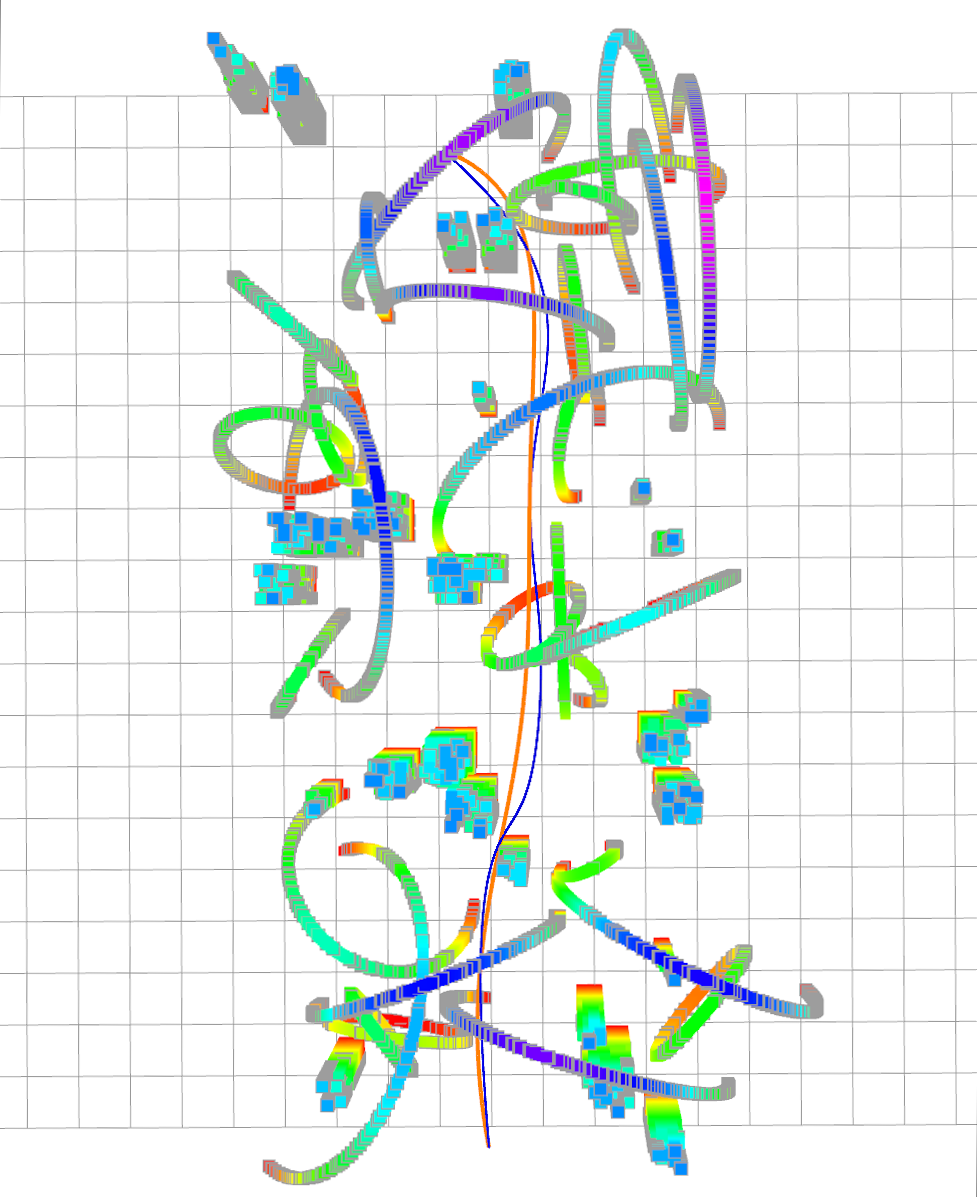}
        }
        \subfigure[Collision case of MINCO trajectory]{
        \includegraphics[width=.22\textwidth]{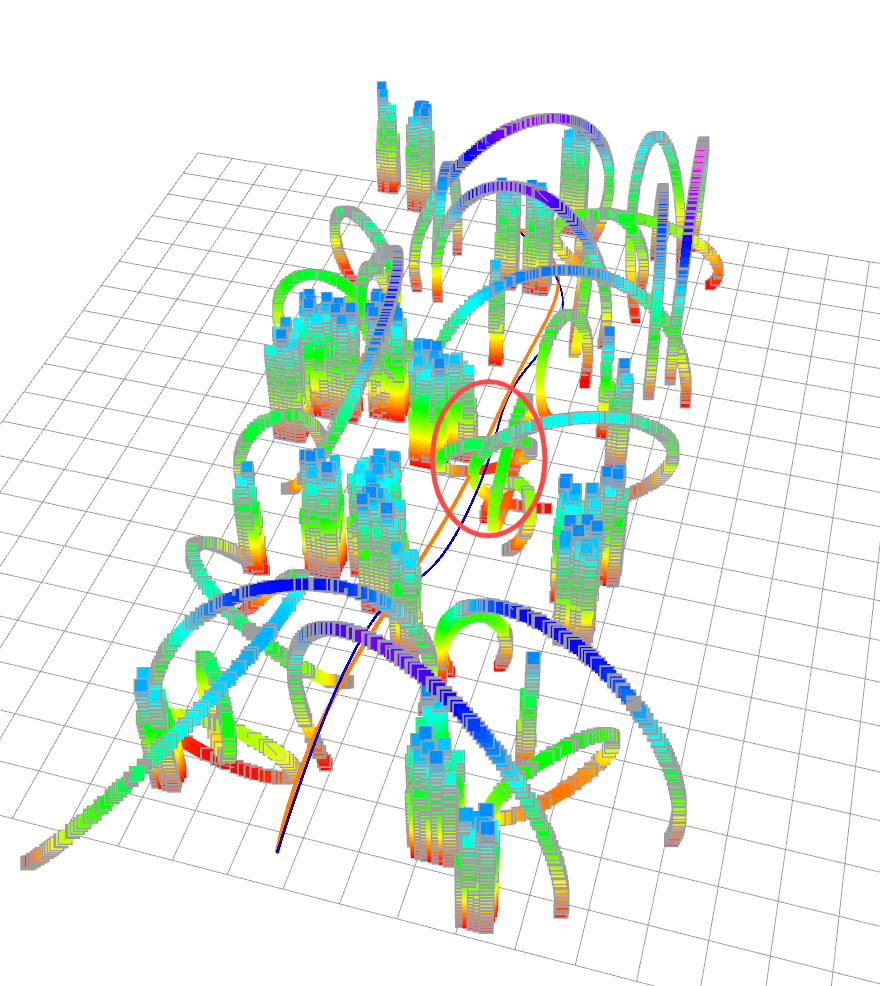}
        }
    \caption{Trajectory Comparision}
\end{figure}

To rigorously validate the trajectory performance, thousands of trajectories with varying mean lengths are generated across multi-scale simulation environments. The quantitative comparative results are systematically summarized in Fig. 6.

\begin{figure}[htbp]
    \centering
    \includegraphics[width=1.0\linewidth]{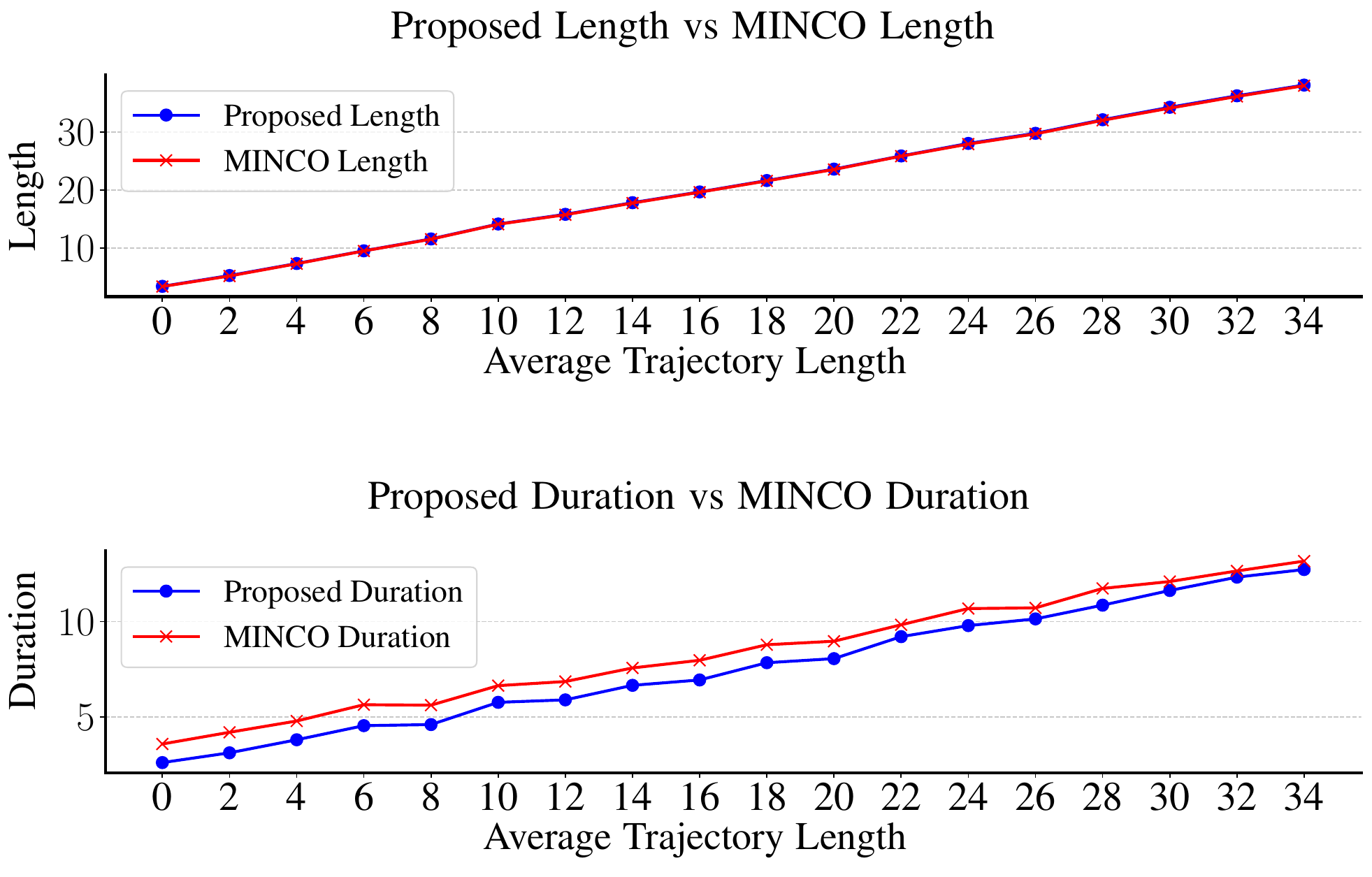}
    \caption{Comparison under different map sizes}
\end{figure}

The results shown in Fig. 6. demonstrate that in environments of the same scale, our method achieves shorter flight durations while maintaining trajectory lengths consistent with those of MINCO method.

To accommodate real-world large-scale scenarios, a simulation environment measuring $30\times30\times4 m^3$ was constructed, generating 2,000 trajectories with minimum length of $15m$ for further performance benchmarking. The result is shown in Table I.
As demonstrated by the tabulated results, the proposed method exhibits superior trajectory length and energy consumption metrics under smaller decay factor while achieving improved flight duration when operating with larger decay factor. 

\begin{table}[htbp]
    \centering
    \footnotesize
    \renewcommand{\arraystretch}{1.5}
    \setlength{\tabcolsep}{3pt}
    \caption{Benchmark Comparision}
    \scalebox{1.0}
    {
    \begin{tabular}{@{}ccccc@{}}
        \toprule
        \textbf{Method} & \multicolumn{1}{c}{\begin{tabular}[c]{@{}c@{}}\textbf{Avg Length}\\ $(m)$\end{tabular}} & \multicolumn{1}{c}{\begin{tabular}[c]{@{}c@{}}\textbf{Avg Duration}\\ $(s)$\end{tabular}} & \multicolumn{1}{c}{\begin{tabular}[c]{@{}c@{}}\textbf{Avg Energy}\\ $(m^2/s^6)$\end{tabular}} &  \\ \midrule
        MINCO$^*$ & 25.49 & 9.43 & 3.68  \\
        Proposed ($\gamma=0.3$) & 25.52 & $\mathbf{8.63}$ & 5.7 \\
        Proposed ($\gamma=0.1$) & $\mathbf{25.04}$ & 9.86 & $\mathbf{2.1}$ \\
        \bottomrule
    \end{tabular}
    }
\end{table}

\begin{itemize}
\item MINCO$^*$ \cite{wang2022geometrically}: Jerk energy minimization is achieved via either fixed total time or linear-time regularization.
\end{itemize}

Comparative analysis of constraint violations from identical simulation datasets (Table II) reveals that the proposed method exhibits superior adherence to safety corridor constraints when compared with the MINCO method. Particularly under smaller decay factors, complete constraint satisfaction of maximum velocity and acceleration bounds was observed. These experimental findings substantiate that our iterative optimization framework achieves better constraint compliance while maintaining satisfactory performance metrics.

\begin{table}[htbp]
    \centering
    \footnotesize
    \renewcommand{\arraystretch}{1.5}
    \setlength{\tabcolsep}{3pt}
    \caption{Constraint Violation Length Ratio}
    \scalebox{1.0}{
    \begin{tabular}{cccc}
    \toprule
        \textbf{Method} & \textbf{SFC}  & \textbf{Velocity}  & \textbf{Acceleration} \\ \midrule
        MINCO$^*$ & 2.6$\%$ & 0.94$\%$ & $\backslash$ \\
        Proposed ($\gamma=0.3$) & 0.45$\%$ & 0.6$\%$ & $\mathbf{0.0\%}$ \\
        Proposed ($\gamma=0.1$) & $\mathbf{0.3\%}$ & $\mathbf{0.0\%}$ & $\mathbf{0.0\%}$ \\
    \bottomrule
    \end{tabular}
    }
\end{table}

\subsection{Real World Experiment} 
The experimental platform is equipped with a 12th Gen Intel® Core™ i5-1240P CPU and a RealSense D455 depth camera. In this hardware configuration, the proposed trajectory optimizer demonstrates real-time capability, operating at an average frequency of 50 Hz, which is sufficient for the majority of motion planning scenarios. During the experiments, we used VINS-Fusion \cite{qin2019b} to provide positioning information and the Nonlinear Model Predictive Control (NMPC) \cite{Falanga2018PAMPCPM} algorithm for trajectory tracking. 

In real-world experiments, multiple flight trials are conducted across various obstacle-laden scenarios as shown in Fig. 7., with comparative analysis performed against MINCO trajectories. 
To ensure the safety during the physical experiments, we set the maximum speed to $0.6 m/s$ and the maximum acceleration to $2 m/s^2$ uniformly.  

Under equivalent constraints on maximum velocity and acceleration parameters, trajectories generated by the proposed method exhibit a $13\%$ reduction in average flight time compared to those produced by the MINCO method. Our method demonstrated superior performance in both
safety metrics and temporal allocation across most scenario. Results of the physical experiments can be found in our attached video. 

\begin{figure}[htbp]
    \centering
    \includegraphics[width=6cm,height=8.9cm]{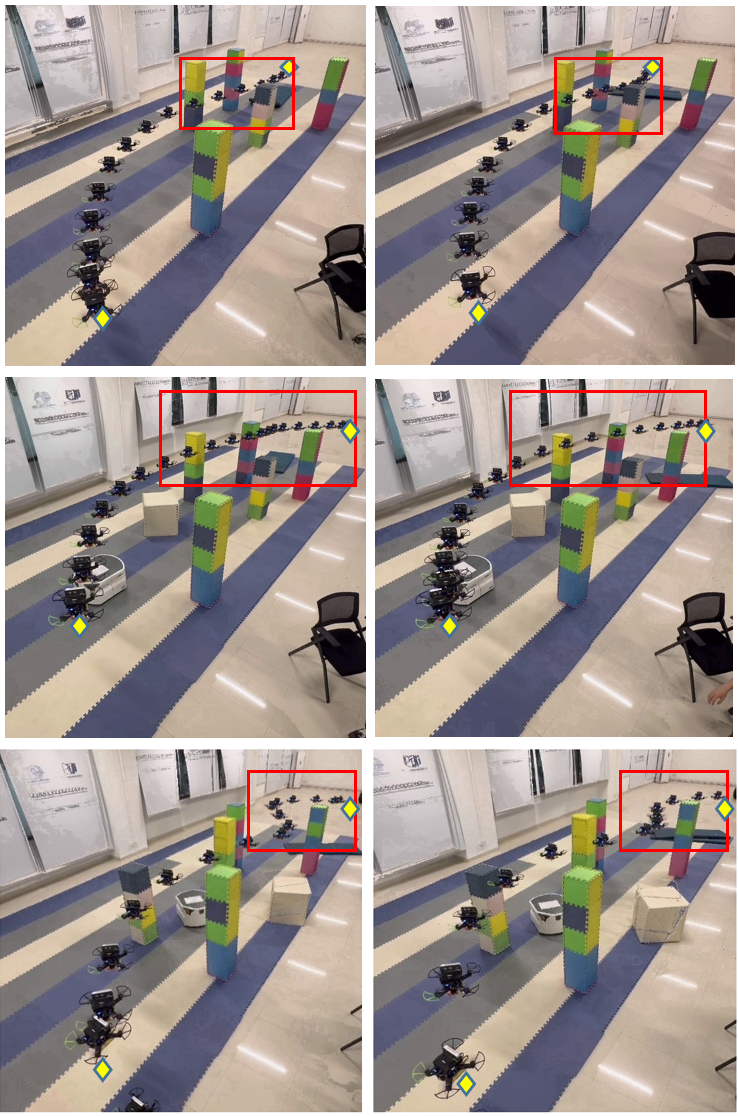}
    \caption{MINCO (left) and Proposed (right), Yellow markers denote the start/end positions, while red bounding boxes show trajectory segments where the proposed method demonstrates superior performance over MINCO method. }
\end{figure}

\section{CONCLUSIONS}
In this paper, a novel spatial-temporal iterative optimization framework for UAV trajectory generation is proposed. The optimization framework has three key advancements: spatial-temporal decoupling which simplifies optimization complexity, convex-hull-based constraints that ensure continuous collision avoidance, and guidance gradient integration instructs convergence of the optimization. Experimental validation confirms superior dynamic feasibility and constraint adherence compared to state-of-the-art methods. 

\addtolength{\textheight}{-12cm}   




\bibliographystyle{IEEEtran}
\bibliography{IEEEabrv, reference}


\end{document}